\documentclass{article}

\usepackage{microtype}
\usepackage{graphicx}
\usepackage{subfigure}
\usepackage{booktabs} 
\usepackage{amsmath}
\usepackage{amssymb}
\usepackage{amsfonts}
\usepackage{enumitem}
\usepackage{titlesec}
\usepackage{lipsum}
\usepackage{siunitx}
\usepackage{bbm}
\usepackage{bm}

\DeclareMathOperator*{\argmax}{argmax}
\usepackage{tikz}
\newcommand*\circled[1]{\tikz[baseline=(char.base)]{
            \node[shape=circle,draw,inner sep=1pt] (char) {\scriptsize \textsf{#1}};}}

\titleformat{\subsection}[runin]{}{}{}{}[]

\usepackage{hyperref}

\hypersetup{draft}

\usepackage[accepted]{icml2019}

\icmltitlerunning{Synthetic Returns for Long-Term Credit Assignment}

\newcommand{\david}[1] {{\color{blue} {}}}

\newcommand{\sam}[1] {{\color{blue} {}}}

\renewcommand{\subsubsection}[1] {{\color{blue} {}}}

\begin{document}

\setlength{\abovedisplayskip}{2pt}
\setlength{\belowdisplayskip}{10pt}

\interfootnotelinepenalty=10000

\twocolumn[

\icmltitle{Synthetic Returns for Long-Term Credit Assignment}



\icmlsetsymbol{equal}{*}

\begin{icmlauthorlist}
\icmlauthor{David Raposo}{equal,dm}
\icmlauthor{Sam Ritter}{equal,dm}
\icmlauthor{Adam Santoro}{dm}
\icmlauthor{Greg Wayne}{dm}
\icmlauthor{Theophane Weber}{dm}
\icmlauthor{Matt Botvinick}{dm}
\icmlauthor{Hado van Hasselt}{dm}
\icmlauthor{Francis Song}{dm}
\end{icmlauthorlist}

\icmlaffiliation{dm}{DeepMind, London, UK}

\icmlcorrespondingauthor{Sam Ritter}{ritters@google.com}

\icmlkeywords{Machine Learning, ICML}

\vskip 0.3in
]




\printAffiliationsAndNotice{\icmlEqualContribution} 

\begin{abstract}

Since the earliest days of reinforcement learning, the workhorse method for assigning credit to actions over time has been temporal-difference (TD) learning, which propagates credit backward timestep-by-timestep. This approach suffers when delays between actions and rewards are long and when intervening unrelated events contribute variance to long-term returns. We propose \textit{state-associative} (SA) learning, where the agent learns associations between states and arbitrarily distant future rewards, then propagates credit directly between the two.
In this work, we use SA-learning to model the contribution of past states to the current reward. With this model we can predict each state's contribution to the far future, a quantity we call ``synthetic returns''. TD-learning can then be applied to select actions that maximize these synthetic returns (SRs).
We demonstrate the effectiveness of augmenting agents with SRs across a range of tasks on which TD-learning alone fails. We show that the learned SRs are interpretable: they spike for states that occur after critical actions are taken. Finally, we show that our IMPALA-based SR agent solves Atari Skiing -- a game with a lengthy reward delay that posed a major hurdle to deep-RL agents -- 25 times faster than the published state-of-the-art.

\end{abstract}

\section{Introduction}
Bellman's seminal work on optimizing behavior over time established an approach to assigning credit to actions in a sequence which remains the default today: give an action credit in proportion to all of the rewards that followed it \citep{bellman1957}. This approach continued in the era of reinforcement learning (RL), especially in the form of temporal-difference (TD-) learning \citep{sutton2018reinforcement}, due in large part to TD's mathematical tractability and its compelling connection with neuroscience \citep{schultz1997neural}. Today, TD-learning remains the workhorse of cutting-edge deep-RL agents; but, as deep-RL is applied to increasing challenging tasks, the cracks in TD-learning are beginning to show \citep{badia2020agent57}.


The problem with this classic approach to credit assignment is that it provides no mechanism to ``skip over'' unrelated events (including rewards) that intervene between actions and the rewards that come as consequence of those actions. This leads to unpredictable variance in the return signal that can prevent value function approximators from representing, and policies from optimizing, crucial future rewards \citep{hung2019optimizing,arjona2018rudder,mesnard2020counterfactual,harutyunyan2019hindsight}.

\sam{Francis asks "whether the difficulty with TD credit assignment is due to TD itself or the fact that we use it with neural network function approximators, finite unroll lengths, etc...?". The correct answer is option 1. Try to make this unambiguous in the next draft.}

\begin{figure}[t]
\begin{center}
\centerline{\includegraphics[width=\columnwidth]{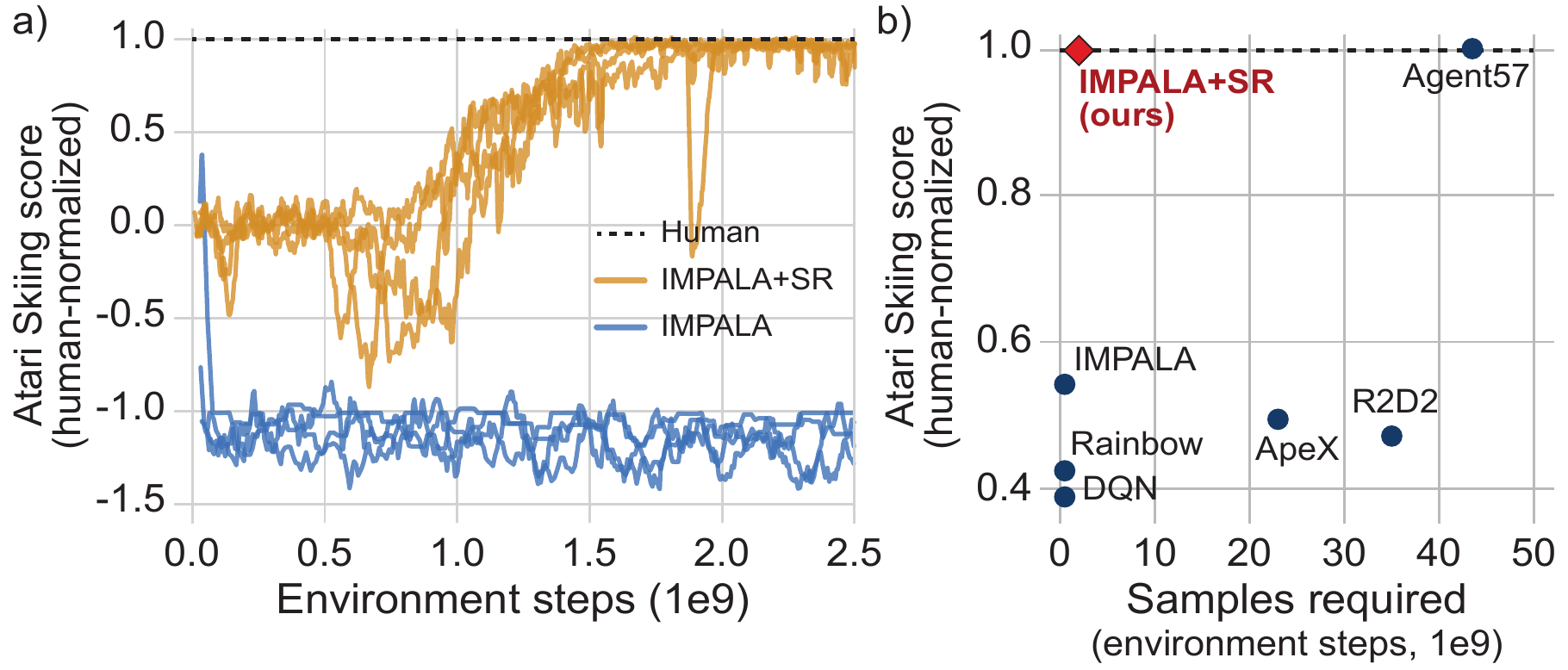}}
\caption{Agent performance on \textit{Atari Skiing} and comparison with past agents. 
Because of its delayed reward structure, Skiing was one of the last games in the Atari Suite to be solved by deep-RL. Our IMPALA-based Synthetic Returns agent obtains human performance, while its counterpart without SRs fails to learn. Moreover, the SR-augmented agent solves the task using 25 times fewer environment steps than the previously published state-of-the-art set by Agent57 \citep{badia2020agent57}. (a) Learning curves for SR-augmented agent (IMPALA+SR) and ablation without SRs (IMPALA). Showing four best seeds per agent. Dotted line indicates human performance. (b) Performance versus the number of samples required to obtain that performance, measured by the total number of environment steps, for each agent's best seed reported in the literature \cite{mnih2015human, hessel2018rainbow, horgan2018distributed, espeholt2018impala, kapturowski2018recurrent,badia2020agent57, badia2020never}.}
\label{atari_skiing_curves}
\end{center}
\end{figure}

\begin{figure*}[ht]
\includegraphics[width=\textwidth]{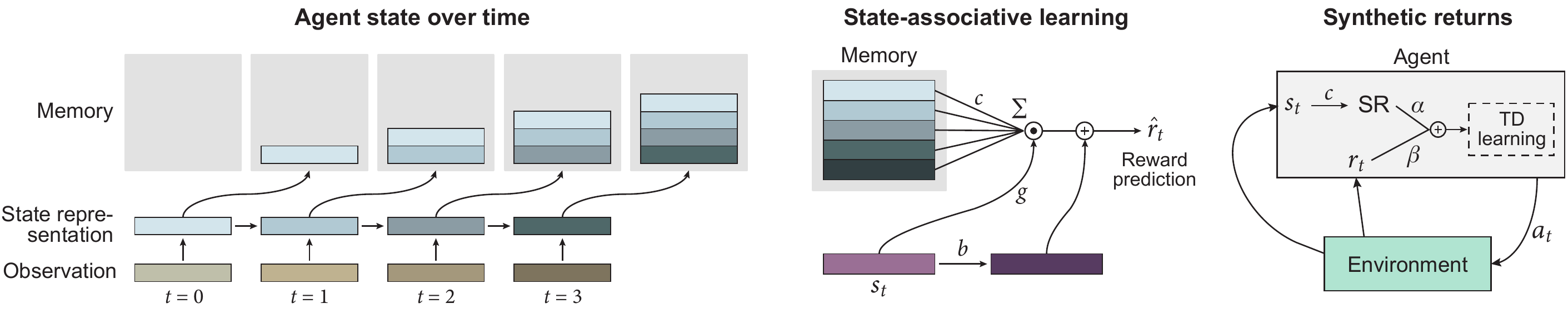}
\caption{(Left) A depiction of the data structure over which we perform state-associative learning. An agent produces a state representation from the observation it receives on each timestep. This can be done using an LSTM, for example, to encode the recent past. Further, the agent maintains an episodic memory containing one state representation for each timestep so far in the current episode. (Center) State-associative learning. $\hat{r}_t$ is learned to predict the current timestep's reward. Crucially, $\hat{r}_t$ is comprised of a baseline term $b(s_t)$ designed to capture the utility of the current state representation $s_t$ plus a sum over a shared ``memory contribution function'' $c(\cdot)$ designed to reflect the difference in the current returns given the presence of a particular past state representation. (Right) A depiction of the use of the memory contribution function as a ``synthetic return'', which is added to the environment's reward ($r_t$) for the current timestep and optimized for by TD-learning.}
\label{method_figure}
\end{figure*}

In this work we propose state-associative (SA) learning, an alternative approach to credit assignment designed to overcome these limitations. In SA-learning, the agent learns associations between pairs of states where the earlier state is predictive of reward in the later state.
These associations allow the agent to assign credit directly from the later state to the previous state, skipping over any timesteps between.

We integrate this approach with current deep-RL agents by using a model of rewards (learned via SA-learning) to ``synthesize'' returns at the current timestep. These ``synthetic returns'' estimate the reward that will be realized at some arbitrary point in the future given the current state. We then use standard TD to learn a policy that maximizes these synthetic returns.

We show that an off-the-shelf deep-RL agent \citep[IMPALA, ][]{espeholt2018impala} augmented with synthetic returns (SRs) is able to solve a range of credit-assignment tasks it is otherwise unable to solve.
Further, we show that the SR model learns to predict the importance of states in which critical actions are taken. This demonstrates that SA-learning is effective at assigning credit to a state even when that state and the future reward it predicts are separated by long delays and unrelated rewards.

Finally, we demonstrate the effectiveness of our approach in \textit{Atari Skiing}, a game that poses a long-term credit assignment challenge which represents a major impediment for deep-RL agents (see \citet{badia2020agent57} for discussion). We show that
an IMPALA agent augmented with SRs is able to solve Atari Skiing while IMPALA without SRs fails.
Further, our SR-augmented agent reaches human performance with 25 times fewer environment steps than the previously published state-of-the-art set by Agent57\cite{badia2020agent57}.

\sam{hopefully by this point the reader sees that SA is different from but compatible/complementary with TD-learning.}

\sam{would be good to make clear that this paper is the "first step", introducing something very novel. so don't expect us to be solving everything across the board.}

\section{Method}
\label{method}


The goal of SA-learning is to discover associations between pairs of states, where the agent's occupancy of the earlier state in the pair is predictive of the later state's reward. After discovering such an association, SA-learning assigns credit for the predicted reward \textit{directly} to the earlier state, ``skipping over'' any intervening events. Consider for example the \textit{Key-to-Door} task, proposed by \citet{hung2019optimizing}, wherein an agent has an opportunity to pick up a key, then experiences an unrelated task with its own rewards, and later has to open a door to receive reward. Crucially, opening the door is contingent on having picked up the key. Notice that in this task, whether or not the agent picks up the key is predictive of whether the agent will receive reward when it attempts to open the door. As we will show, SA-learning takes advantage of this fact to assign credit directly to the ``key-pickup'' event. 

This contrasts with classic credit assignment approaches. To reinforce the picking up of the key, these methods require the learning of a value function, or the training of a policy, using a target that includes all of the intervening, unrelated rewards. It is sensitivity to this irrelevant variance that SA-learning avoids.

The key contribution of our work is to show how SA-learning can be done efficiently in a deep-RL setting. Our primary insight is that we can use a buffer of state representations---one for each state so-far in the episode\footnote{This data structure has a long history in deep-RL, and is sometimes called ``episodic memory'' \citep{oh2016control, wayne2018unsupervised, ritter2018been, fortunato2019generalization}.}---to learn a function that estimates the reward that each state predicts for some arbitrary future state. Our method learns reward-predictive state associations in a \textit{backward-looking} manner; it uses past state representations stored in the buffer to predict the current timestep's reward. We show that these learned reward-predictive associations can then be used in a \textit{forward-looking} manner to reinforce  actions that lead to future reward. We do so by applying the learned function to the current state, then using the function's output as an auxiliary reward signal.
 
We now present a method that encapsulates this process of SA-learning as a module that can be added to standard deep-RL agents to boost their performance in tasks that require long-term credit assignment\footnote{We will release an implementation of our method as a self contained, framework agnostic, JAX module \citep{deepmind2020jax}, that can be added on to a typical RL agent.}. Our module is designed to work with an agent that learns from \textit{unrolls} of experience \citep{mnih2015human,mnih2016asynchronous,espeholt2018impala,kapturowski2018recurrent}, which may be as short as a single timestep, and will generally be much shorter than the episode length\footnote{For a formal definition of ``episode'' and related concepts in RL, see \citet{sutton2018reinforcement}.}. We assume the agent uses these unrolls to compute gradient updates for neural networks that perform critical functions---such as value estimation and policy computation.
We assume that the agent has an internal state representation that may capture multiple timesteps of experience---for example, an LSTM state \citep{hochreiter1997long} or the output of a convolutional network.

Our module makes two additions to this standard agent. First, it augments the agent's state to include a buffer of all of the agent's state representations for each timestep so far in the episode. Second, it adds another neural network which is trained in the same optimization step as the others. The new network is trained to output the reward for each timestep $t$ given the state representations $\{s_0, ..., s_t\}$ in the augmented agent state. This can be achieved via the following loss:

\begin{align}
\mathcal{L}_{c,g,b} = \Big \| r_t -
g(s_t)\sum_{k=0}^{t-1} c(s_{k}) - b(s_t) \Big \|^2
\label{basic_arch}
\end{align}

where $g(s_t)$ is a sigmoid gate whose output is in the range $[0, 1]$, and $b(s)$ and $c(s)$ are neural networks that output scalar real values. 
After training this network, we interpret $c(s_t)$ as a proxy for the future reward attributable to the agent's presence in state $s_t$, \textit{regardless of how far in the future that reward occurs}. We compute this quantity for each state the agent enters, and use it to augment the agent's reward:

\begin{align}
\label{reward_summation}
\tilde{r}_t = \alpha \, c(s_t) + \beta \, r_t
\end{align}

where $\alpha$ and $\beta$ are hyperparameters, and $r_t$ is the usual environment reward at time $t$. 

To motivate this interpretation of $c(s_t)$, we will explain the network's architecture by breaking it down into its components. The summation over $c(s_k)$ is the core of the architecture. In essence this summation casts reward prediction as a \textit{linear regression problem}, where variance in the reward signal must be captured by a sum of weights, with each weight corresponding to one past state. After fitting this model, we can interpret the weight $c(s_k)$ as the amount of reward that the agent's presence in $s_k$ ``contributed'' to some future state $s_t$. For a more formal treatment of this interpretation, see Supplement Sections \ref{theory} and \ref{additive_regression_limitation}.

Notice that $c(s_k)$ does not need to receive the $s_t$ as an input. This is a crucial design feature: it allows us to query the contribution function for an arbitrary state without having access to the future state it will contribute to. In practice, this means that we can use $c(\cdot)$ as a \textit{forward-looking} model to predict, at time $t$, the rewards that will occur at a future time $\tau$ regardless of whether $s_\tau$ is in the same unroll as $s_t$. In other words, $c(\cdot)$ allows us to ``synthesize'' rewards during training before they are realized in the environment. We will refer to the output of $c(\cdot)$ as a ``synthetic return''.

For this to be possible, we need the \textit{backward-looking} function (the summation term) to determine to which future states the memory contributions are relevant. This is the role of the gate $g(s_t)$. Notice that in using this gate, we are assuming that the contributions from past states to future states is sparse---i.e., that there is a single future state for which the contributions of past states is non-zero. This is an assumption about the \textit{structure of the environment} which holds for many tasks. Future work will be needed to generalize our approach to tasks that do not satisfy this assumption.

Finally, notice that we introduce a separate function $b(s_t)$ to compute the contribution of the current state to the current reward, instead of using the shared function $c(s_t)$.
This is intended to encourage the model to prefer to use the current state to predict the current reward by allowing more expressivity in the function approximator\footnote{We could enforce this preference by first using the current state predict the current reward, then using the memories to predict the residual, as described in Section \ref{atari_skiing_text}. However, for all of our tasks except Atari Skiing we found that the agent performed equally well without this enforcement, so we left it out for the sake of simplicity.}. The benefit of this design choice is that $c(s_k)$ can be interpreted as an \textit{advantage}, estimating how much better---or worse---things will be in the future state, due to the addition of the current state to the trajectory. This is especially useful in environments where important events lead to non-positive rewards (see Section \ref{key_to_door} and Suppl. Figure \ref{key_door_appendix} for details). In early experiments we found that adding the separate function $b(s_t)$ worked better than using $c(.)$ for the current state, but we did not systematically compare these two variants.

Note that the $\alpha$ and $\beta$ hyperparameters in Eq. \ref{reward_summation} trade off between two different learning objectives: the discounted total of future reward, and the predicted reward at an arbitrary future state. We chose to retain the influence of the discounted total reward objective because the inductive bias it embodies is effective for many tasks; temporal recency is often an effective heuristic for assigning credit. This is especially true of modern deep-RL benchmarks, which have become popular in part because they are amenable to the available RL algorithms, which optimize for discounted total reward. In such tasks, SA requires time and data to learn a model that associates temporally nearby states. In contrast, this association of nearby states is \textit{built-in} to classic RL algorithms. We opt for the best of both worlds: let the TD algorithm handle short timescales by optimizing for discounted total reward, while SA handles the longer timescales that are difficult for TD.

\section{Experiments}
\label{experiments}

In the following experiments we apply SA-learning to produce synthetic returns for IMPALA \citep{espeholt2018impala}, a distributed advantage actor-critic agent. Our agent implementation is built in Tensorflow \citep{abadi2016tensorflow}. With the exception of the experiments in Section \ref{atari_skiing_text}, we used convolutional neural network outputs as the representations of past and current states for SA-learning and an LSTM for the policy (see Suppl. Section \ref{suppl:architecture} for other architecture details).

\subsection{Chain Task}

\begin{figure}[t]
\includegraphics[width=\columnwidth]{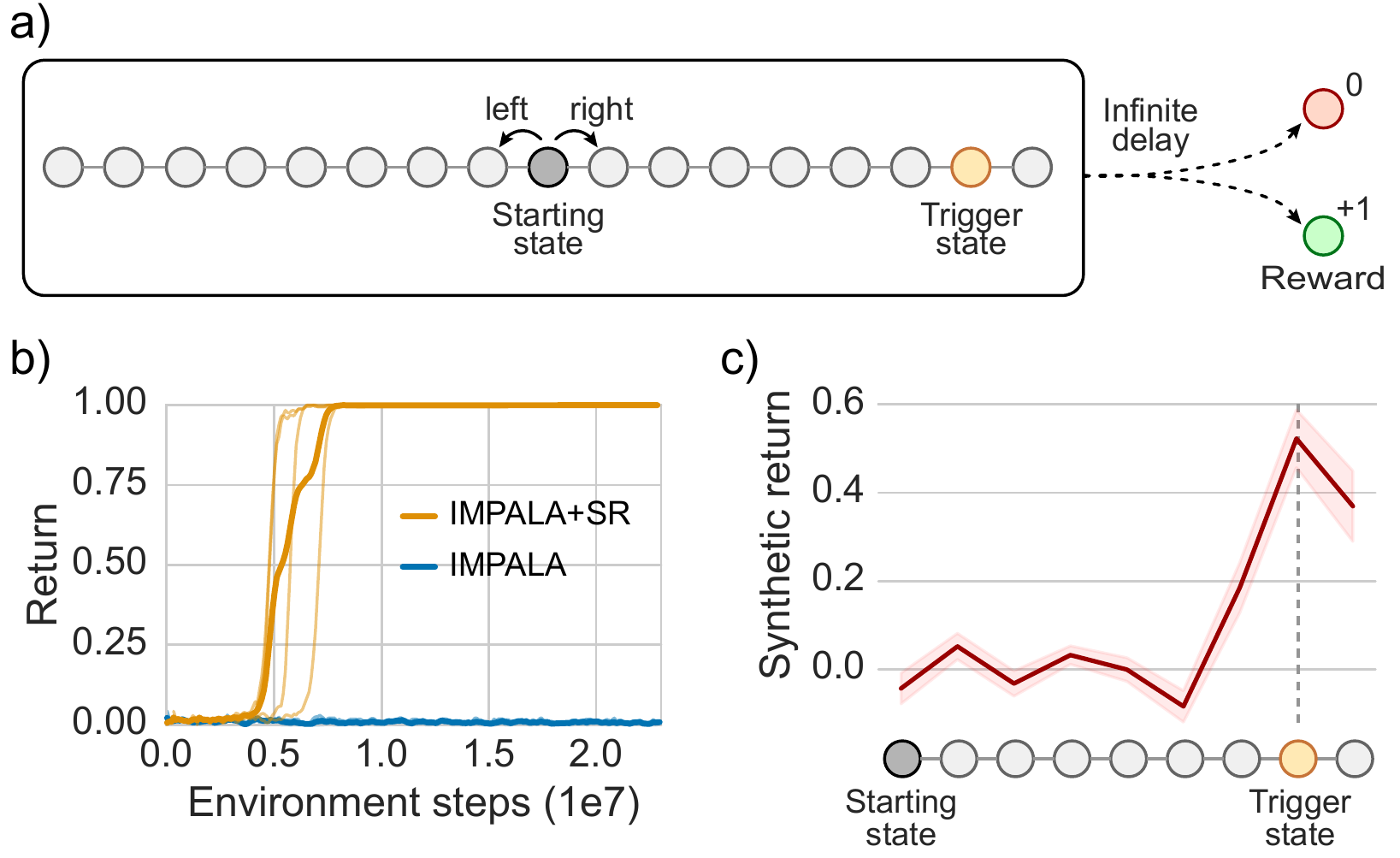}
\caption{
(a) Schematic of the Chain task. The task consists of a chain of 17 states. Starting from the state exactly in the center (dark gray circle), the agent navigates the chain by choosing "left" or "right" actions. After a fixed number of steps (10) the agent will transition to a state that gives a reward of +1 (green circle), but only if the agent visited a specific state in the chain, which we call the \textit{trigger state}. If the agent did not visit the trigger state, the agent will instead transition to a non-rewarding state (red circle). We simulate an infinite delay during the transition to the green or red states. We do so by blocking the Bellman backups (i.e. setting the TD discount factor to zero) for the step corresponding to this transition. (b) Learning curves for agents with (orange) and without (blue) synthetic returns. The plots show four seeds for each condition (thin lines) and mean across seeds (thick lines). (c) The magnitude of the synthetic return for each one of the states in the chain to the right of the starting state. Red line indicates average across episodes (between $3e6$ and $1e7$ training steps); shaded pink area indicates standard deviation. \david{this should perhaps be across seeds rather than episodes.} Dashed gray line indicates the trigger state.
}
\label{chain_task}
\end{figure}

To build an intuition for the approach, we first demonstrate its performance in the Chain task, a minimal and interpretable long-term credit assignment task (depicted in Figure \ref{chain_task}a). The task consists of a chain of states (hence the name), with two possible actions: one moves the agent one state to the right, the other moves the agent one state to the left. At the beginning of an episode, the agent is placed in the state exactly in the center of the chain: the starting state. The agent is then allowed to move freely along the chain for a fixed number of steps, which we set to 10 in our experiments. After this number of steps the agent transitions to one of two possible states: one which is rewarding (+1), and another one which is non-rewarding. Transitioning to the rewarding state is determined by whether the agent visited a particular state---the \textit{trigger state}---in any of the previous 10 steps. Critically, we simulate an infinite temporal delay during the transition to the rewarding or the non-rewarding state by blocking the Bellman backups for the step corresponding to this transition. We achieve this by simply setting the discount factor used in the TD learning algorithm to zero for that step.

In our experiments, the trigger state location was fixed across training, seven steps to the right of the starting state. This results in a 0.8\% chance of visiting the trigger state in an episode with a uniform random policy. A state's observation is a unique one-hot vector of length 18.

\textbf{The SR-augmented agent solves the task reliably}, while the baseline agent without SRs fails (Figure \ref{chain_task}b). This indicates that SA-learning allows the agent to assign credit from the rewarding state to the actions that led the agent to visit the trigger state, in spite of the fact that no returns propagate between those steps via Bellman backups. Note that it is expected that the baseline agent---which learns via TD only---should fail at this task, because there is no way for it to assign credit the actions that brought the agent to the rewarding state.

\textbf{The learned synthetic returns (SRs) spike at the trigger state} (Figure \ref{chain_task}c). The agent should be incentivized to visit the trigger state in order to collect reward later on. This is in fact what the SRs learn to signal to the agent, effectively replacing the return signal that would be provided by the Bellman backups: the SRs take a large value for the trigger state and a lower value (close to zero) for the other states.

\sam{it might also be good to say something about when during training this effect emerges, and whether it changes as training progresses (e.g. does it decay?). edit: i guess this is covered in the delayed-reward catch figure. maybe it's worth referring to that briefly when we first show SR spikes in the chain section.}

\subsection{Catch with Delayed Rewards}

\textit{Catch} is a very simple, easy to implement game that has become a popular sanity-check task for RL research. In this task the agent controls a paddle located at the bottom of the screen and, moving right or left, attempts to ``catch'' a dropping ball. The ball drops in a straight line, starting from a random location on the top of the screen. For each ball that the agent catches with the paddle a reward of +1 is received. An episode consists of a fixed number of runs---i.e. catch attempts. The goal of the agent is simply to maximize the number of successful catches per episode. In our implementation of the task the screen is a $7\!\times\!7$ grid of pixels, and the paddle and the ball are each represented with a single pixel.


This version of the task (which we call \textit{standard Catch}) can trivially be solved by most RL agents. However, introducing a simple modification that delays all the rewards to the end of the episode makes it substantially more difficult, or even unsolvable, for our current best agents. We will call this variant of the task \textit{Catch with delayed rewards}. To give a concrete example, if in an episode the agent makes seven successful catches out of 20 runs, it will not receive any reward until the very last step of the episode, at which time point it will receive a reward of +7.

\begin{figure}[t]
\includegraphics[width=\columnwidth]{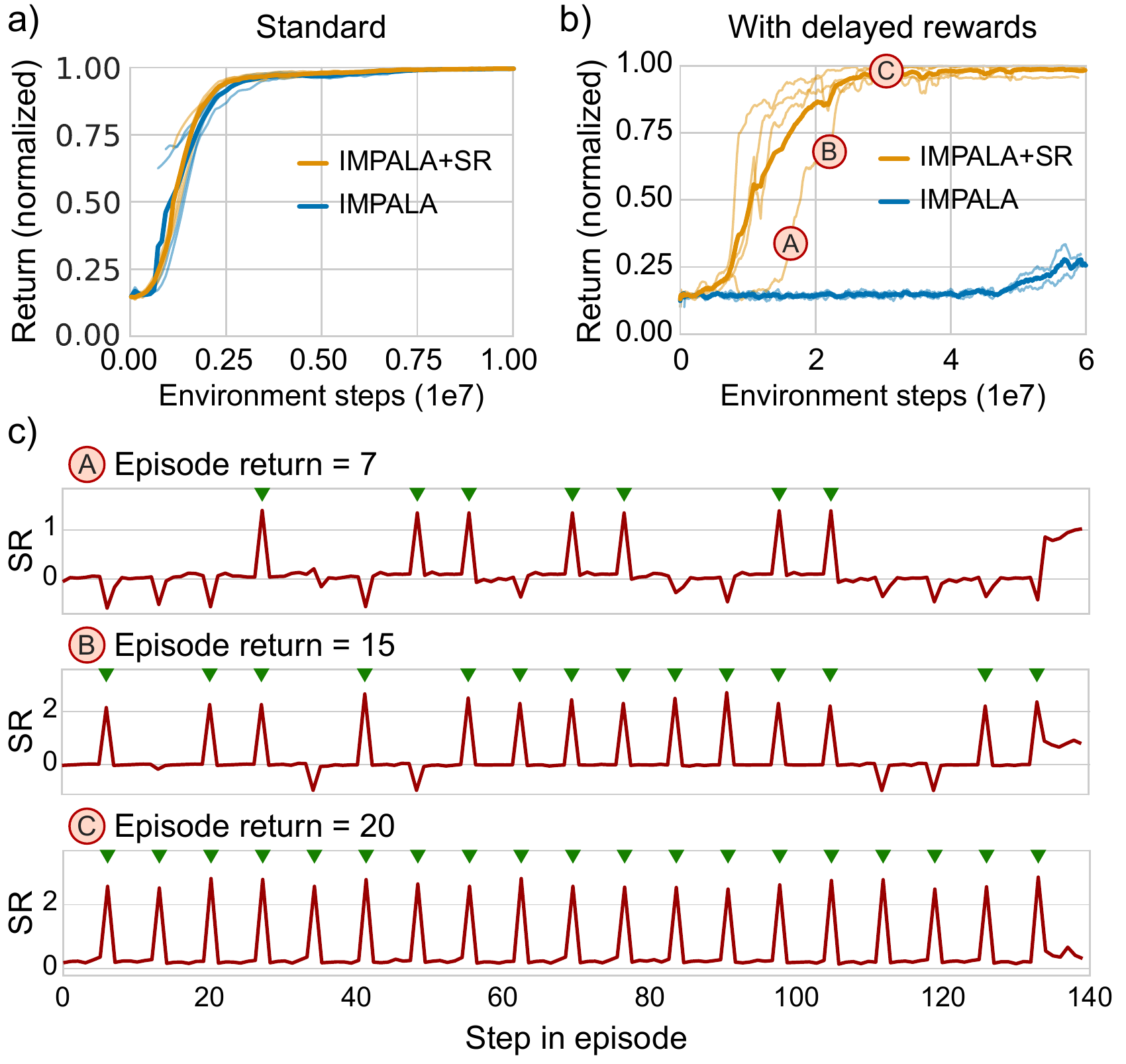}
\caption{Results on Catch. (a) Learning curves on the standard Catch. Both the baseline IMPALA and IMPALA with SRs succeed in learning to solve this task. The performance of the two agent is indistinguishable. (b) Learning curves on \textit{Catch with delayed rewards}. Only the agent with SRs was able to consistently solve this variant of the task. The plots show four seeds for each condition (thin lines) and mean across seeds (thick lines). (c) The magnitude of the synthetic returns for three example episodes, taken from early (A), mid (B) and late (C) training. Green triangles indicate the time steps in which a successful catch occurred. This analysis reveals that the spikes in synthetic return coincide with successful catches, providing an extra learning signal to the agent. Small dips in synthetic return coincide with missed catches.}
\label{mini_skiing}
\end{figure}

\begin{figure*}[t]
\center
\includegraphics[width=.8\textwidth]{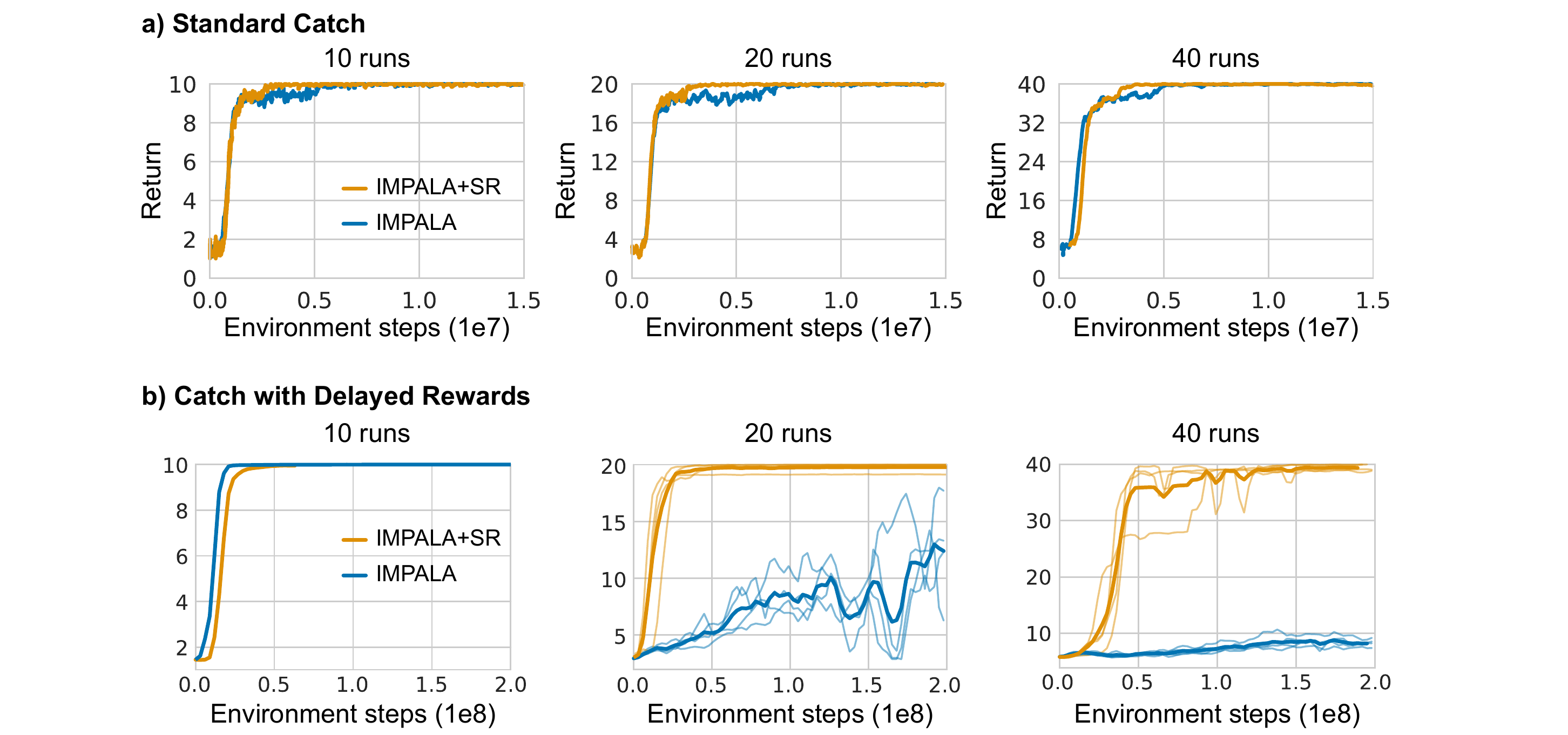}
\caption{Learning curves for \textit{standard Catch} (a) and \textit{Catch with delayed rewards} (b) for an increasing number of runs per episode (10, 20, 40). Both the agent augmented with SRs (IMPALA+SR) and the agent without SRs (IMPALA) perform well on \textit{standard Catch}. Their performance and sample efficiency are maintained at the same level as the number of runs per episode increases. With delayed rewards, the performance of the baseline agent degrades considerably as we increase the number of runs per episode. The SR agent is able to achieve high performance across number of runs, with a small impact in its sample efficiency.}
\label{mini_skiing_appendix}
\end{figure*}

Unsurprisingly, our experiments show that the baseline IMPALA agent can efficiently solve \textit{standard Catch}. Our SR-augmented IMPALA agent also performs very well on this version, achieving a performance that is indistinguishable from that of the original agent (Figure \ref{mini_skiing}a). On \textit{Catch with delayed rewards}, only the SR-augmented agent learned to solve the task perfectly after $6e7$ environment steps, consistently receiving the maximum reward of 20 at the end of each episode. The baseline agent (without SRs) performed significantly worse, achieving a mean performance of 13 out of 20 across four seeds after $2e8$ environment steps (see Suppl. Figure \ref{mini_skiing_appendix}b, middle).

The analysis in Figure \ref{mini_skiing}c shows that as an example learning curve begins to climb (time point labeled \circled{A} in Figure \ref{mini_skiing}b), the \textbf{SR spikes at time steps on which the agent made a successful catch}. The SR also shows small dips that are aligned with missed catches. As the curve climbs to fully solve the task (time points labeled \circled{B} and \circled{C}), the SR function continues to spike for the increasing number of successful catches. We can infer from this analysis that the SR is, as we hoped, acting in place of a reward signal to reinforce the actions that lead to successful catches and discourage the ones that lead to misses.

The baseline agent's best performance on the delayed-reward variant of the task required a discount factor close to one ($0.99$). This is in contrast with the SR agent that solved the task with the same discount factor both agents used to solve the standard version of the task ($0.90$). This suggests that, in the absence of an immediate reward from the environment, the agent can make use of the synthetic returns to signal a successful event and learn which actions to take \textit{locally} (temporally speaking) to that event.


\begin{figure*}[ht]
\includegraphics[width=\textwidth]{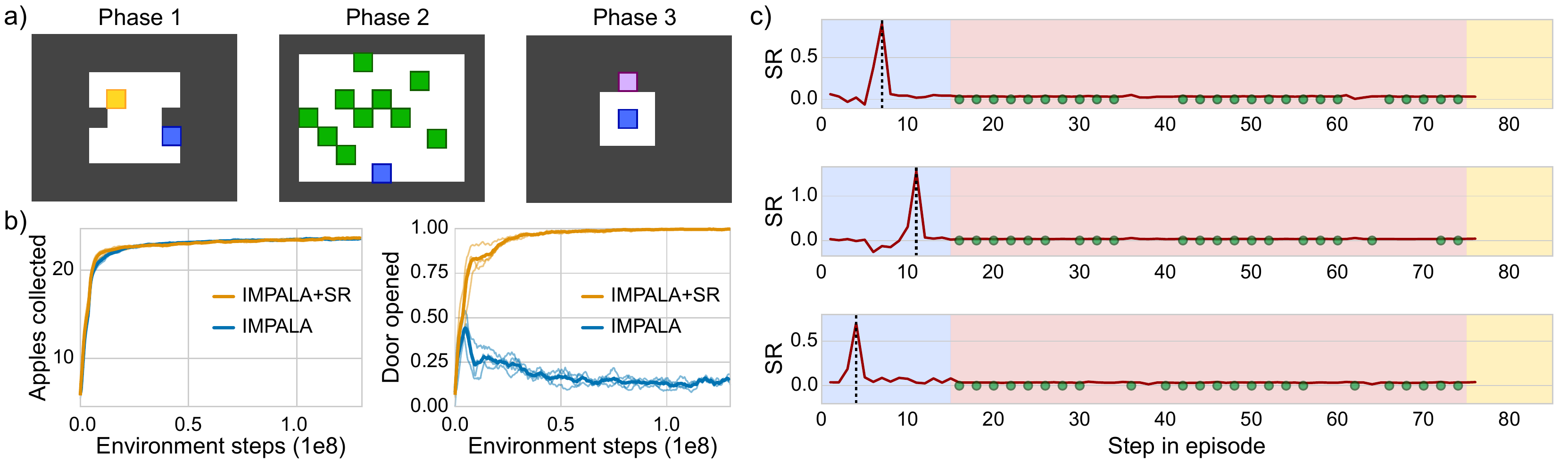}
\caption{Key-to-Door task and results. (a) The three phases of the task. In phase one, the agent (blue pixel) can navigate to the key (yellow pixel) and pick it up. This phase has a fixed duration of 15 time steps. The key's location and the agent's starting location are randomly chosen in each episode. In the second phase, the agent can collect apples (green pixels), which are randomly placed, to receive a reward of +1 for each. This phase has a fixed duration of 60 time steps. In the third phase, the agent can open the door (purple pixel) and collect +5 reward, but only if it has picked up the key in the first phase. The episode terminates when the agent opens the door or after 10 time steps. (b) Learning curves for an Impala agent with (orange) and without (blue) synthetic returns. The plot shows four seeds per condition (thin lines) and mean across seeds (thick lines). Both the baseline agent and the agent with SRs learn to collect all the available apples (left). In contrast, only the agent with SRs learns to consistently collect all the apples and open the door in phase three (right). (c) The synthetic returns magnitude for each time step of three example episodes. Blue, pink and yellow shaded areas delineate the three different phases; green circles indicate the collecting of an apple; dotted line indicates the time step in which the agent collected the key. This analysis reveals that the spike in the synthetic returns during phase one coincides with the agent collecting the key, which can be used as an extra learning signal.} 
\label{key_door}
\end{figure*}

Because the rewards in \textit{standard Catch} are presented immediately upon a successful catch, the performance of an agent is not significantly affected by the number of runs per episode (see Suppl. Figure \ref{mini_skiing_appendix}a). In contrast, with delayed rewards the task becomes harder for a typical agent as the number of runs per episode increases. The reason for this is that longer episodes impose a longer delay between a valuable action and its outcome. Moreover, because the total reward is received at the end of the episode as an aggregate of all the successful catches, the contribution of a single catch to the final reward gets smaller as we increase the number of runs per episode.
Our results show that the baseline agent can only perform well on this version of the task when the number of runs per episode is low. As we increase the number of runs, the agent's performance quickly degrades. This is not the case for the SR agent. The agent is able to perfectly solve the task with a much greater number of runs per episode with minimal impact to its sample efficiency (see Suppl. Figure \ref{mini_skiing_appendix}b).

The results so far suggest that augmenting RL agents with SRs enables them to solve long-term credit assignment problems. They also provide some confidence that our algorithm does not have an impact on an agent's performance in tasks that do not require long-term credit assignment, such as \textit{standard Catch}. This observation is further supported by our experiments on \textit{Atari Pong} (see Section \ref{sec:pong}).

\subsection{Key-to-Door}
\label{key_to_door}

The previous results demonstrated the ability of SA-learning to learn over a long delay. However, long delays are only part of the problem: the real challenge occurs when distracting events, and especially unrelated rewards, occur during the delay. To investigate whether our SA-learning agent can solve such problems, we trained our agents on a well-established task from the credit assignment literature: \textit{Key-to-Door} \citep{hung2019optimizing}. We developed a grid world version of the task using the Pycolab engine \citep{stepleton2017pycolab}. The Key-to-Door task proceeds in three phases. The first phase consists of a room with one key that the agent can pick up. Doing so yields no reward. Both key and agent are placed randomly in the room. After 15 time steps, the agent automatically transitions to the second phase which takes place in another room. In this room there are ``apples'', randomly scattered, which yield rewards when picked up (+1) by the agent. The second phase has a fixed duration of 60 time steps. In the third and final phase, the agent can open a door and receive reward (+5), but only if it picked up the key in the first phase. The episode terminates after 10 time steps or immediately after the agent opens the door.

The second phase constitutes a \textit{distractor task}, which emits rewards that interfere with learning the relationship between picking up the key in the first phase and receiving reward in the third phase. For this reason---and contrary to what was the case for \textit{Catch with delayed rewards}---using a high discount factor here is not generally favourable, even to an agent which can only rely on TD-learning.

Our results show that the baseline IMPALA agent has no trouble learning to pick up the apples, which emit immediate rewards, during the second phase (Figure \ref{key_door}b, left). However, this agent fails to learn to consistently pick up the key in the first phase in order to open the door in the third phase (Figure \ref{key_door}b, right). These results replicate the typical agent's failure reported by \citet{hung2019optimizing}. The SR-augmented agent, on the other hand, learns to systematically pick up the key and open the door, while still collecting all the available apples during the distractor phase.

By inspecting the synthetic returns magnitude over the course of an episode, we observe a consistent spike during the first phase that is aligned with the moment the agent picks up the key (Figure \ref{key_door}c). Throughout the second and third phases, the SRs keep mostly flat and close to zero, even with the occurrence of a mixture of rewarding (collecting an apple, opening the door) and non-rewarding states (navigating). From the point of view of the synthetic returns, these states are inconsequential.

\begin{figure*}[t]
\center
\includegraphics[width=.8\textwidth]{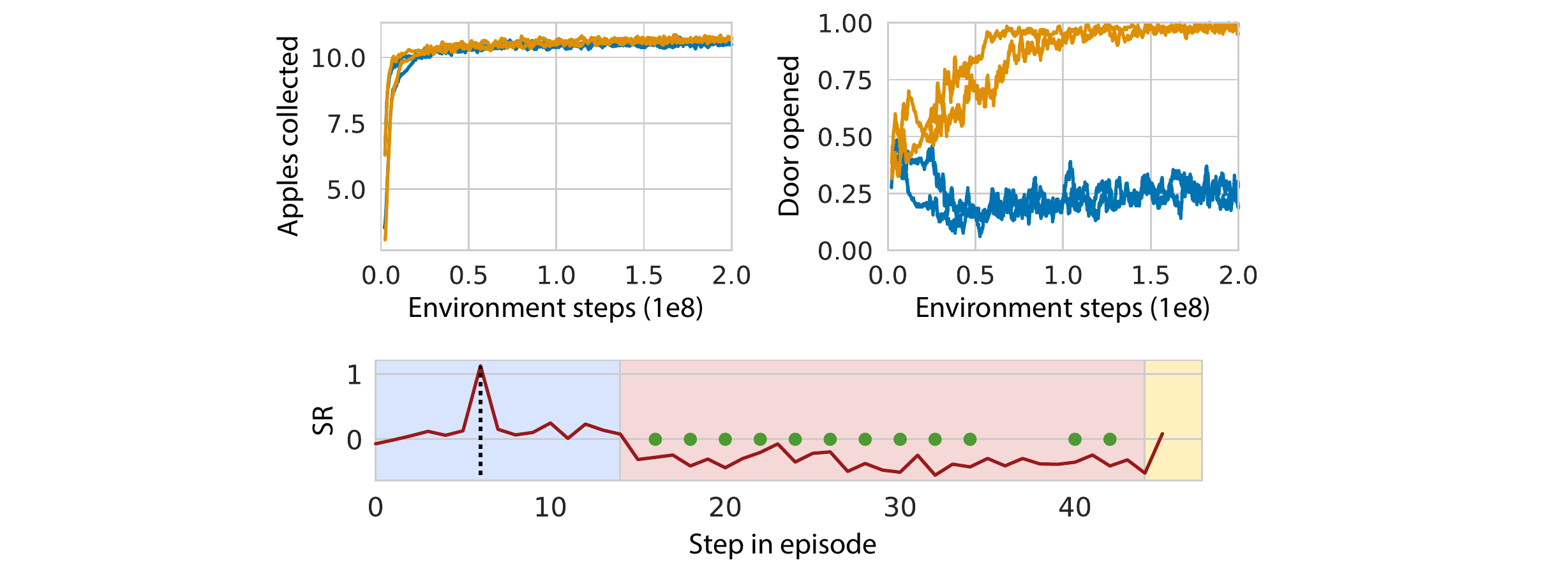}
\caption{\textit{Key-to-Door} task variant. Like in the original task, in this variant the agent can pick up a key in phase one in order to open a door in phase three. The difference is that in this version opening the door results in zero reward and there is a penalty per step of --1 during phase three. Phase two remained the same: the agent can pick up apples that give a reward of +1. In these experiments we made this phase shorter for faster turnaround (30 time steps, versus 60 in the original version).}
\label{key_door_appendix}
\end{figure*}

We ran experiments on two other variants of the Key-to-Door task. In the first one, we modified the third phase such that the reward for opening the door with the key was zero, while adding a penalty of --1 per step. The other phases remained unchanged. The SR-augmented agent successfully solved this task as well. Inspecting the SR magnitudes over the course of an episode, we still observe a positive spike aligned with the moment of collecting the key, even though the reward for opening the door was now zero (see Suppl. Figure \ref{key_door_appendix}). This illustrates a feature of our algorithm that we described before: the SRs learn to measure the utility of a state for an arbitrarily far future state in the form of an \textit{advantage} relative to the expected return in that future state.

In the second variant, we introduced two keys (yellow and red) that the agent could collect during the first phase. Opening the door in the third phase with the yellow or the red key led to a reward of --1 or --2, respectively. Additionally, the agent received penalty of --5 for not opening the door. The SR agent was also able to solve this task successfully, learning to consistently collect the yellow key.

\subsection{Pong}
\label{sec:pong}

We trained our agents in the Atari game \textit{Pong} to test whether the synthetic returns negatively impacted performance in a task that does not require long-term credit assignment.

In Pong the agent controls a paddle, located on the left side of the screen, by moving it up or down. A second paddle, located on the right, is controlled by the environment simulator (the opponent). The goal is to bounce the ball back and forth until the opponent misses the ball, which results in a reward of +1. If the agent misses a ball, a reward of --1 is received. An episode ends when the agent's score reaches either --21 or +21.

\begin{figure}[h]
\begin{center}
\centerline{\includegraphics[width=.8\columnwidth]{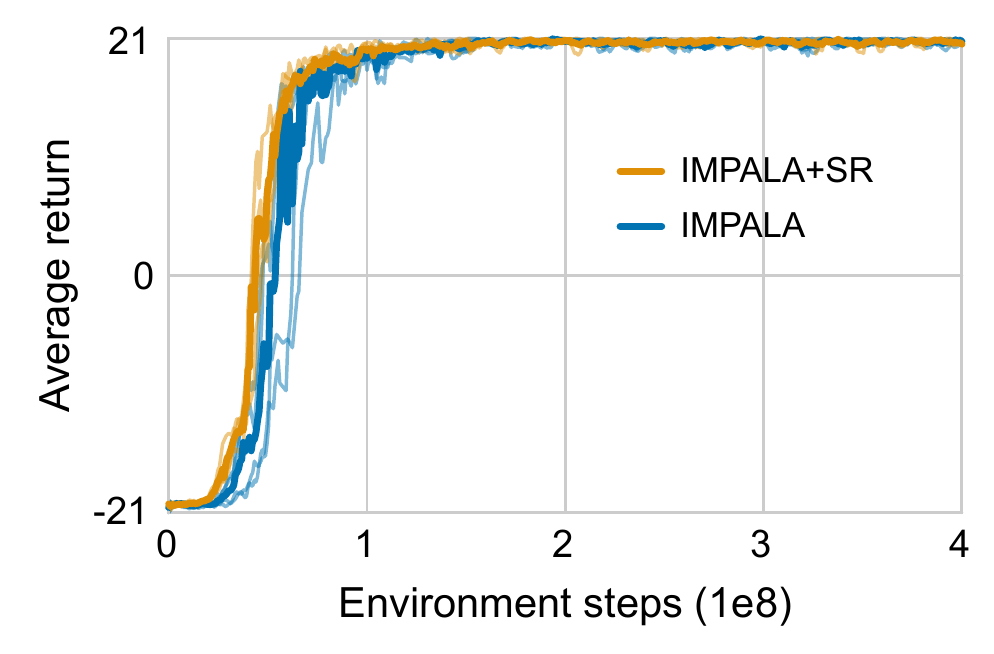}}
\caption{Learning curves for \textit{Atari Pong}. Plot shows four seeds per agent (thin lines) and mean across seeds (thick lines). Results demonstrate that augmenting an agent with SRs does not disrupt its performance in tasks that do not require long-term credit assignment. The agent with SRs (IMPALA+SR) matches the one without SRs (IMPALA) in performance and sample efficiency.}
\label{pong}
\end{center}
\end{figure}

Figure \ref{pong} shows the results on this task. We observed that the SR-augmented agent performs identically to the IMPALA baseline. This result provides extra evidence that the SR algorithm does not sacrifice generality in order to solve long-term credit assignment tasks.

\subsection{Atari Skiing}
\label{atari_skiing_text}


In \textit{Skiing} the agent is tasked with hitting as many gates as possible while avoiding obstacles.
Similarly to \textit{Catch with delayed rewards}, this task poses a hard credit assignment challenge due to the long delay between actions and their outcomes: the reward---or, in this case, the penalty for each gate missed---is only received in aggregate at the end of an episode.

For this task we forced the memory baseline to predict as much variance as possible, by separating the loss into two parts:

\begin{align}
\begin{split}
  \| r_t - b(s_t)\|^2  \\
  \Big \| r_t - \textrm{stopgrad}(b(s_t)) - \sum_{k=0}^{t-1} g(s_t) \, c(s_k) \Big \|^2  \\
\end{split}
\end{align}

With this approach we observed better results than when we used the single loss from Eq. \ref{basic_arch}. This multi-stage loss enforces the \textit{memory advantage} interpretation discussed in Section \ref{method}. We used the single-stage loss in the other tasks because in those tasks it achieved the same performance and is simpler.

We used convolution neural network outputs as the representations of past states $s_k$, and an LSTM hidden layer as the representation of the current state $s_t$.  We report results for $\beta$ set to zero (see Eq. 2), which we found to work best in this task. This means that in this task, SR-augmented agent optimized only synthetic returns, and ignored the environment rewards. The hyperparameter sweep we used is described in Supplement Section \ref{as_hparams}.

Figure \ref{atari_skiing_curves} shows the perfomance of the baseline IMPALA versus IMPALA with SRs. The SR agent solves the task in less than 1.5 billion steps. In contrast, the baseline IMPALA fails to learn, in line with previous results \citep{espeholt2018impala}. Agent57 \citep{badia2020agent57} was to our knowledge the first deep-RL agent to achieve human performance on this task. That agent's best seed required 43.5 billion steps to reach human performance. \textbf{The SR-augmented IMPALA thus achieves a $\bm{25\!\times}$ sample efficiency improvement over the published state-of-the-art}.

\section{Related work}

\citet{hung2019optimizing} introduced a mechanism for credit assignment, called Temporal Value Transport (TVT), whereby agents store states in an episodic memory and then use the ability to recall specific memories as a way to directly assign credit to the states associated with those memories. The credit assigned to a previous state corresponds to the return estimated at the timestep at which that state is recalled. The estimated return, or value, associated with the later state is simply “transported back” to the past state.

TVT does not learn a model of rewards. In other words, it does not learn to decompose the rewards into components to be explained by different past states. This prevents TVT from being applicable to problems with states that only partially predict some distant future reward. This is the case in \textit{Catch with delayed rewards}, for example.

Another undesirable consequence of this is that the agent has to experience the state from which reward can be predicted in order for value to be transported back to a previous state. Mechanistically speaking, both the past state and the future state---respectively, the receiver and the sender of the transported value---need to occur within the same unroll. This makes TVT difficult or impossible to apply in practice to problems involving very long delays, like \textit{Atari Skiing}. Our approach instead uses SA-learning to predict value potentially far beyond the end of the unroll.

RUDDER \citep{arjona2018rudder} trains an LSTM to predict total episode reward at every timestep. RUDDER then uses the differences between the LSTM's successive predictions as a reward signal for a standard RL algorithm. Like our method, this achieves a decomposition of reward into contributions from each timestep in the episode. Our method is in principle less expensive to compute because it does not require a sequential unroll to learn the decomposition. 
Further, our method \textit{decomposes each timestep's reward} instead of the total episode reward, such that our decomposition can be learned using standard unrolls. This ease of adoption along with superior scalability are advantages of our approach.

Recent work has developed methods for more effective credit assignment using hindsight and counterfactual estimation \citep{harutyunyan2019hindsight, mesnard2020counterfactual,buesing2018woulda}. 
Other recent work has used attention weights as a proxy for credit:  \citet{ferret2019self} apply this approach to improve transfer, while \citep{liu2019sequence} focus on artificial delayed-reward problems using sequence-wide prediction.

\section{Discussion}
\label{discussion}

The present work represents an important first step for state-associative learning (SA) in deep-RL. However, there are currently a number of limitations that we hope to address in the future. First, we expect the performance of the gated SA architecture to suffer when the environment is not sparse. Second, the current method is not sensitive to the number of times a predicted state occurs. Third, because of our use of a multiplicative gate, we cannot offer convergence guarantees as to the semantics of $c(s_t)$. Additionally, because of our use of an additive regression model, we cannot offer a rigorous guarantee of optimal credit assignment, and indeed when multiple states predict the same reward, which state ``gets credit'' is unconstrained. For more detail on all of these points, see Section \ref{method} and the Supplement. In Suppl. Sections \ref{theory} and \ref{additive_regression_limitation} we discuss SA-learning variants that may overcome these drawbacks, and we look forward to future work providing theoretical and empirical analysis of these variants.

Finally, in our Atari Skiing experiments our best seeds perform extremely well, but we still observe substantial seed variance and hyperparameter sensitivity (see Suppl. Figure \ref{atari_skiing_full_sweep}). In future work we will explore ways of improving robustness.

In this work, we showed that SA-learning with IMPALA is able to achieve superior sample efficiency to Agent57 in a challenging benchmark task with delayed reward. This is particularly striking, because Agent57 benefits from sophisticated methods for exploration and online hyperparameter tuning that IMPALA does not have. By combining SA-learning with a more recent agent, such as Agent57, more significant gains in state-of-the-art performance and sample-efficiency may be possible. 



\section*{Acknowledgements}
We thank Anna Harutyunyan for insightful feedback and discussion of our writeup and formal aspects of the work; Pablo Sprechmann, Adri{\`a} Puigdom{\`e}nech Badia, and Steven Kapturowski for discussion of the approach and experiments, as well as access to Agent57 learning curves; Daniel Wynham for discussion about the Atari Skiing environment; and Daan Wierstra and Tim Scholtes for discussion about the architecture and experiments. We thank everyone at DeepMind whose work on the company's technical and organizational infrastructure made this work possible.

\bibliography{bibliography}
\bibliographystyle{icml2019}

\section{Supplement}

\subsection{Formalism: State-Associative Learning}
\label{theory}

In this section we provide a more formal treatment of SA-learning and the algorithm we used in our experiments. The goal of SA-learning is to infer the utility of a given state. SA-learning shares this goal with TD-learning, but differs in its definition of utility. In TD algorithms, utility (or value) is defined as the sum of discounted rewards,

\begin{align}
V^{\pi}(s_t) = \mathbb{E}_{\pi} \sum_{\tau=t+1}^T \gamma^{\tau-(t+1)} r_\tau
\end{align}

Here $s_t$ denotes the agent state at time $t$, $\pi$ denotes the agent's policy, $r_\tau$ denotes the reward the agent receives on the $\tau^{th}$ timestep, and $\gamma$ denotes the discount factor. Note that we here consider the episodic case, where $T$ is a random variable that denotes the timestep on which the episode terminates.

Consider that this value formulation includes all reward that came after $t$, regardless of whether or not the events at time $t$ had any influence on those rewards. Although by definition the irrelevant rewards will have a mean of zero, when training on samples of experience these rewards act as noise in the regression target for any function approximator $V_{\pi}(s_t)$ attempting to estimate the value function. This noise can deter or entirely thwart value learning.

\sam{Is this the best/only motivation for SA over TD?}

\sam{One other difference: SA value functions also have spikes instead of plateaus (assuming you use the normal variants). This is very tricky to think about and trickier to discuss, because whether you get this effect depends on which variant of proper SA you use, none of which we actually have in our code. So maybe it's best to leave this for now. Also its not clear to me what/whether there's any advantage to having spikes instead of plateaus.}

We propose to mitigate this problem by formulating a utility function $D^{\pi}(s_t)$ made up only of rewards that are predicted by the agent's presence in state $s_t$

\begin{align}
D^{\pi}(s_t) = \mathbb{E}_{\pi}
\sum^T_{\tau=t+1}
\mathcal{A}(s_t, s_\tau) \, r_\tau
\label{utility_function}
\end{align}

where ${\mathcal{A}}(s_t, s_\tau)$ is a function that specifies the proportion of $r_\tau$ which is predicted by $s_t$. In essence, ${\mathcal{A}}$ is intended to capture the reward-predictive association between an earlier state $s_t$ and a later state $s_\tau$.

The product $\mathcal{A}(s_t, s_\tau)\,r_\tau$ is the absolute amount of $r_\tau$ that $s_t$ predicts. In a sense, this is the amount of reward that $s_t$ ``contributes'' at $s_\tau$. We will refer to this product as the \textit{contribution function} $C(s_t, s_\tau)$.

How can we estimate the utility function $D^{\pi}(s_t)$? First, we propose to estimate the contribution function $C$ via a form of linear regression,

\begin{align}
\mathcal{L}_{c} = \Big \| r_t - \sum_{k=0}^{t} c(s_k, s_t) \Big \|^2
\label{regression_model}
\end{align}

where $c(s_k, s_t)$ are regression weights that estimate $C(s_k, s_t)$\footnotemark. Notice that $\tau$ is used to index states after time $t$, while $k$ is used to index states before $t$. 

In finding $c(s_k, s_t)$, the model learns to associate pairs of states where the earlier state is predictive of the later state's reward, hence the name state-associative learning. When our agent state contains a buffer of the current episode's state representations (memory, Figure \ref{method}, left) we can compute this model's output using the short experience unrolls that are used in typical deep-RL frameworks. 

\footnotetext{When multiple $s_{k<t}$ are predictive of the same variance in $r_t$, this additive regression model does not constrain which of those states should ``get credit''. See Supplement Section \ref{additive_regression_limitation} for further discussion.}

We can then use the contents of the memory at the end of each episode to compute $\sum_{\tau=t+1}^{T}c(s_t, s_\tau)$. Using episodes sampled during standard agent training, we can fit a neural network $d(s_t)$ to estimate the expectation of that sum

\begin{align}
\hat{D}^\pi(s_t) = \mathbb{E}_{\pi}
\sum_{\tau=t+1}^{T}c(s_t, s_\tau)
\label{D_hat}
\end{align}

thereby achieving an estimate of the utility function in Eq. \ref{utility_function}. We could then use the output of $d(s_t)$ as a synthetic return estimating the utility of $s_t$.

Typical distributed deep-RL training frameworks send unrolls from actors to learners, and do not allow for special logic to apply to the end of episodes, as would be required in order to compute $\sum_{\tau=t+1}^{T}c(s_t, s_\tau)$. Although it would be straightforward to implement a framework that computes this, in this work we elected to develop a method that would integrate nearly effortlessly with the frameworks most researchers are already using.

In order to do that, we assume that the set of associations is sparse; that is, that $c(s, s')$ is non-zero for only one $s'$. This assumption allows us to split $c(s, s')$ into two functions,  $c(s)$ and $g(s')$, each one a function of a single state (see Eq. \ref{basic_arch}). $g(s')$ is a gate between zero and one that learns \emph{whether} past states are predictive of reward at $s'$. $c(s)$ then estimates \textit{how much} reward $s$ contributes to $s'$. 

To gain an intution for the potential benefit of this approach, consider that happens in the idealized case where the learned gate outputs exactly 0 or 1 for all pairs $(s, s')$ , and outputs 1 for a unique $s'$ for each $s$. In this setting  we have a utility function that reflects the reward predicted for a single future state


\begin{align}
\tilde{D}^\pi(s_t) =
\mathbb{E}_{\pi} \, \frac{1}{n} \,
c(s_t)
\sum_{\tau=t+1}^{T} \mathbbm{1}_{\{s'\}}(s_\tau)
\end{align}

where $n$ is the number of occurrences of state $s'$ in the sequence $s_{t+1},...,s_T$,
and $\mathbbm{1}$ is the indicator function.
Notice that the summation is cancelled out by the $\frac{1}{n}$, and so
this utility function is not sensitive to the number of times $s'$ occurs in the sequence $s_{t+1},...,s_T$.

While this idealized setting provides intuition for the kinds of solutions this architecture might learn, unfortunately this setting of the gates is not guaranteed to occur, even in environments that satisfy the sparsity assumption. The reason for this is that having both $g$ and $c$ provides an extra degree of freedom, such that their outputs are ill-constrained. We find nonetheless that in practice this architecture works well, and analyzing the learned $c$ outputs shows that they take on sensible values. Future work can experiment with variants of SA-learning which, unlike this variant, offer convergence guarantees for the semantics of the outputs of $c$. 


\subsection{Combining SA-Learning and TD-Learning}
When combining SA with TD by summing the SA utility function with the environment reward we introduce a problem: we might \textit{double count} reward components. If SA ascribes reward to a state that occurs shortly before the reward, then TD will credit the same state with the same reward. In a sense, our current algorithm does not satisfy \textit{reward conservation}: reward can be ``created'' through this double counting effect.
However, we did not find this to be an issue in practice. The reason for this is that using a low TD discount factor, in tasks in which states contribute to future reward over long delays, effectively washes out this double counting.
Other variants of our algorithm satisfy reward conservation, but we leave empirical evaluation and formalization of these to future work.

\subsection{Using SA-learning to Find Policies}
\label{sa_alone}
SA-learning could conceivably be used to learn policies directly, without invoking TD-learning as we do in this work. This could be done by adding actions to the state embeddings described in Sections \ref{method} and \ref{theory}, yielding state-action utility estimates $U(s_t, a_t)$. Then it is straightforward to apply e.g. the greedy policy 

\begin{align}
\underset{a_t}{\argmax}\, U(s_t, a_t). 
\end{align}

We leave the empirical investigation of this approach to future work.

\subsection{Limitation of Additive Regression}
\label{additive_regression_limitation}
After fitting the model from Eq. \ref{regression_model}, we have

\begin{align}
r_t = \sum_{k=0}^{t}c(s_k, s_t) + \eta \,,
\end{align}

where $\eta$ represents variance not captured by the model. Rewriting this, we see that $c(s_y, s_t)$ captures any explainable variance that was not explained by contributions from the other states

\begin{align}
c(s_y, s_t) = r_t - (\sum_{x=0}^{t, x\neq y}c(s_x, s_t) + \eta) \,.
\label{algebra}
\end{align}

Thus $c(s_k, s_t)$ corresponds to the component of $r_t$ that is \textit{best} predicted by $s_k$ (see Eq. \ref{algebra}). ``Best'' here means that the regression model can minimize its loss more by predicting $r_t$ using $s_k$ than it can using any of the other states preceding $s_t$. Unfortunately, it is not clear that this allocation of reward components produces optimal credit assignment. For example, if there is multicollinearity among the states preceding $s_t$---i.e., multiple prior states are predictive of the same components of $r_t$---then the state which receives credit is unconstrained. An approach that may allow a more rigorous treatment of optimal policies, and which would also constrain the solution in the case of multicollinearity, would be to separate the learning into multiple stages. 

\begin{align}
\begin{split}
  r_t \leftarrow c(s_0, s_t), \\
  r_t - c(s_0, s_t) \leftarrow c(s_1, s_t), \\
  r_t - (c(s_0, s_t) + c(s_1, s_t)) \leftarrow c(s_2, s_t), \\
  ... \\
  r_t - \sum_{k=0}^{t-1} c(s_k, s_t)
  \leftarrow c(s_t, s_t). \\
\end{split}
\end{align}

Preferring simplicity of implementation, in this work we experimented with the additive formulation, and found it to work well in practice. We leave formal analysis and empirical examination of the multi-stage variant to future work.

\subsection*{RL setup}

We used an advantage actor-critic setup for all experiments reported in this paper, with V-trace policy correction (IMPALA) as described by \citet{espeholt2018impala}.

The distributed agent consisted of $256$ actors that produced trajectories of experience on CPU, and a single learner running on a Tensor Processing Unit (TPU), learning over mini-batches of actors' experiences provided via a queue.

We used the RMSprop optimizer for training, except for Skiing experiments which used Adam (see Section \ref{as_hparams}). The following table indicates the tuning ranges of the hyperparameter used for all the other tasks:

\begin{center}
 \begin{tabular}{||l c||} 
\hline
Hyperparameter & Ranges \\
\hline\hline
Agent & \\
\hspace{4mm}Mini-batch size & [32, 64] \\
\hspace{4mm}Unroll length & [10, 60] \\
\hspace{4mm}Entropy cost & [1e-3, 1e-2] \\  
\hspace{4mm}Discount & [0.8, 0.99] \\
SA &\\
\hspace{4mm} $\alpha$ & [0.01, 0.5]\\
\hspace{4mm} $\beta$ & 1.0\\
RMSprop & \\
\hspace{4mm}Learning rate & [1e-5, 4e-4] \\
\hspace{4mm}Epsilon & 1e-4 \\
\hspace{4mm}Momentum & 0 \\
\hspace{4mm}Decay & 0.99 \\
\hline
\end{tabular}
\end{center}

\subsection{Architecture details}
\label{suppl:architecture}

We used a 2-D convolutional neural network (ConvNet) to process pixel inputs. For Catch and Key-to-Door, this ConvNet consisted of 2 layers with 32 and 64 output channels, with $2\!\times\!2$ kernels and stride of 1. Additionally, we included a final layer of size 256, followed by a ReLU activation.

For the Atari games Pong and Skiing, which had larger observations, the ConvNet had 3 layers with 32, 64 and 64 output channels, $3\!\times\!3$ kernels and stride of 2.

For the Chain task we did not use a ConvNet, and instead used a single layer of 128 units followed by a ReLU.

The architecture responsible for the SA-learning component used a memory buffer with capacity corresponding to the maximum number of steps per episode in each game. This varied from 10 (Chain task) to 2000 (Skiing).

The synthetic returns function $c$ was a 2 layer MLP with 256 units per layer and ReLU activation functions. The current state contribution $b$ was  another MLP of the same size. The gate $g$ was produced with a 2-layer MLP with 256 units in the first layer and a single unit in the second.

The output of the ConvNet was passed to a LSTM, with 256 hidden units, followed by the policy network. The policy network consisted of a 256-unit layer followed by a ReLU and two separate linear layers to produce policy logits and a baseline.

The baseline agent, against which we compared our agent's performance, consisted of ablating the SA-learning module altogether by simply setting the loss corresponding to Eq. \ref{basic_arch} to zero.

\subsection{Atari Skiing Hyperparameter Sweep}
\label{as_hparams}

The results of the synthetic return agent presented in the main text correspond to the best four seeds out of 40 seeds (top 10\%) in a random search using log uniform distributions with the following ranges:


\begin{center}
 \begin{tabular}{||l c||} 
 \hline
 Hyperparameter & Ranges \\ [0.5ex] 
 \hline\hline
 Agent &\\
 \hspace{4mm}Mini-batch size & 32\\
 \hspace{4mm}Unroll length & 80\\
 \hspace{4mm}Entropy cost & [1.5e-2, 3.5e-2] \\ 
 \hspace{4mm}Discount & 0.99\\
 SA &\\
 \hspace{4mm} $\alpha$ & [9e-3, 1.4e-2] \\
\hspace{4mm} $\beta$ & 0.0\\
 Adam &\\
 \hspace{4mm}Learning rate & [1.25e-5, 4.5e-5] \\ 
 \hspace{4mm}Epsilon & [1.84e-4, 4e-4] \\
 \hspace{4mm}Beta1 & [0.85, 1] \\
 \hspace{4mm}Beta2 & [0.8, 0.9] \\
 \hline
\end{tabular}
\end{center}

\begin{figure*}[t]
\center
\includegraphics[width=.7\textwidth]{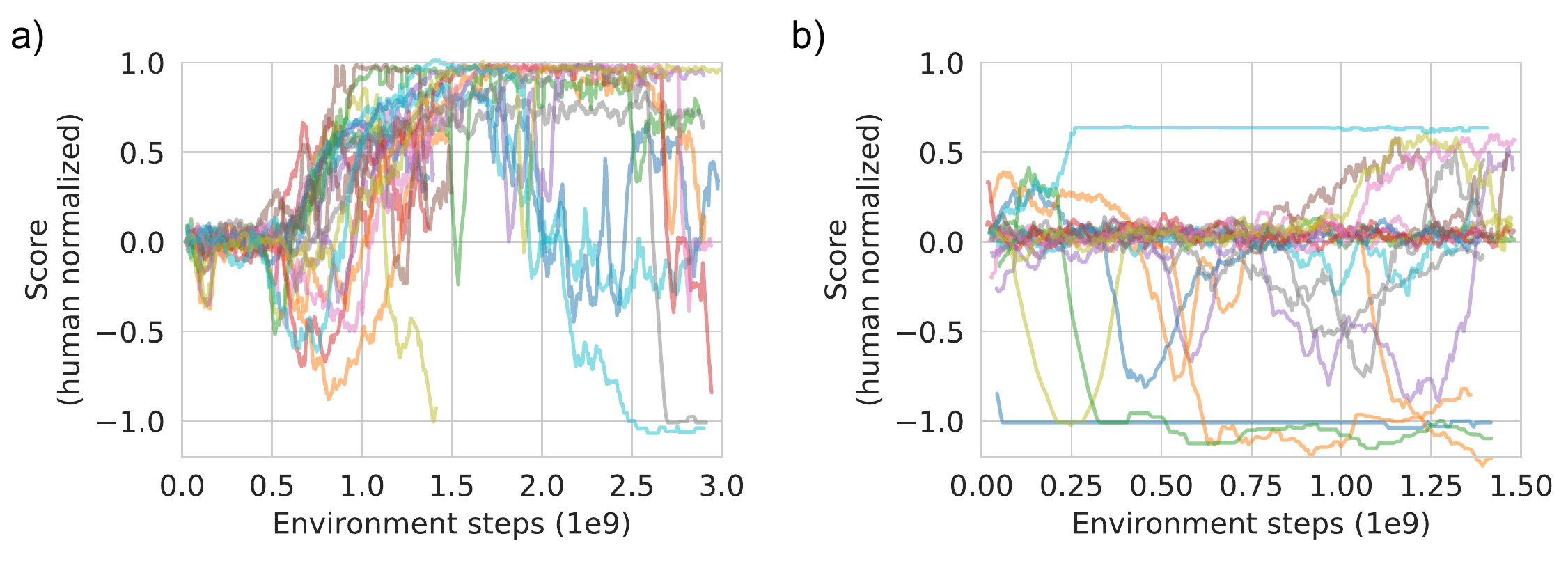}
\caption{Results for all seeds in a random hyperparameter search on Atari Skiing with IMPALA+SR. Training of the less promising seeds was stopped earlier. (a) The best 20 out of 40 seeds. (b) The worst 20 out of 40 seeds. Results indicate seed variance and/or hyperparameter sensitivity. In further analyses (not shown) we observed that there is considerable seed variance within narrow hyperparameter ranges. See Section 5 of the main text for discussion of future research directions for improving robustness.}
\label{atari_skiing_full_sweep}
\end{figure*}

Results for the full sweep are shown in Figure \ref{atari_skiing_full_sweep}.

\end{document}